# Dynamic Motion Modelling for Legged Robots

Mark Edgington, Yohannes Kassahun and Frank Kirchner
Robotics Group
University of Bremen
28359 Bremen, Germany
{edgimar, kassahun, frank.kirchner}@informatik.uni-bremen.de

*Abstract*— An accurate motion model is an important component in modern-day robotic systems, but building such a model for a complex system often requires an appreciable amount of manual effort. In this paper we present a motion model representation, the Dynamic Gaussian Mixture Model (DGMM), that alleviates the need to manually design the form of a motion model, and provides a direct means of incorporating auxiliary sensory data into the model. This representation and its accompanying algorithms are validated experimentally using an 8-legged kinematically complex robot, as well as a standard benchmark dataset. The presented method not only learns the robot's motion model, but also improves the model's accuracy by incorporating information about the terrain surrounding the robot.

## I. INTRODUCTION

A robot motion model, broadly speaking, represents the relationship between commands issued to a robot and the robot movement that results from these commands. In many modern robotic systems, an accurate and reliable motion model serves a crucial role. Most Simultaneous Localization and Mapping (SLAM) approaches, for example, use a motion model in the prediction step of a recursive Bayesian estimation algorithm [1]. An accurate motion model can also be used to perform dead-reckoning, aiding trajectory-following tasks when other forms of odometric information is limited.

In robotics it is a common assumption that a robot's motion model is Markovian. In this case, the motion model is often defined in terms of a probability density function $p(s_t|s_{t-1}, u)$, where $u$ is a command issued to the robot when it is in state $s_{t-1}$. Traditionally, $u$ has been defined (for wheeled robots moving in a 2D plane) as either a set of translational and rotational velocities to be used to directly control the robot's wheels (in which the state $s$ represents the robot pose as a function of time), or as a relative odometry (e.g. a vector $\langle \delta_1, r, \delta_2 \rangle$) where the robot is commanded to turn by an angle, move forward a certain distance, and turn again by another angle (in this case, the robot pose represented by $s$ does not depend on time) [2].

Typically these probabilistic models depend on the existence of a closed-form parameterized deterministic model, consisting of one or more equations that define a future state in terms of the present state and the command issued. By adding uncertainties to this deterministic model, it is transformed into a probabilistic model. For example, a simple deterministic model relating distance to velocity, $d = v\Delta t$, can be made probabilistic by treating $d$ and $v$ as Gaussian-distributed random variables: due to mechanical tolerances in the robot, $v$ has an associated uncertainty that can be modeled using the random variable's $\sigma$ parameter. The dependence on closed-form models is found in nearly all works on this topic up to now. While such models are useful when working with wheeled robots in a plane, they are not adequate for modelling the complex motions of legged robots, for which it is often difficult and time-consuming to find a deterministic mapping from the command-space to the configuration-space.

The motion model representation presented in this paper not only overcomes this limitation, but is also flexible in its ability to incorporate arbitrary sensory information directly into the motion model. Specifically, we present a novel representation and method for online learning of *both the parameters and form* of a robot's motion-model. By allowing terrain information, for example, to be easily incorporated into the model, a more accurate motion-model can be captured, because the movement of a robot is closely tied to the properties of the terrain on which it moves.

This paper is structured as follows: several relevant works are discussed in the next section, after which we formally present our motion-model representation. Following this, an algorithm suitable for dynamically capturing a motion model using this representation is presented. Furthermore, to demonstrate the flexibility of the representation, we show how sensed terrain information can be incorporated into a motion model. Experimental results are then reported which validate the usefulness of the methods and representation. Lastly, conclusions and possible future directions to explore are presented.

## II. REVIEW OF WORKS

Before the appearance of probabilistic SLAM methods, getting an accurate estimate of a robot's position often relied on a parameterized model relating data reported by odometry sensors to the estimated motion of the robot. The problem of robot calibration, which involves properly selecting the parameters of this model, has been the topic of various works. For example, the work of Borenstein and Feng [3] presents methods for manually choosing these parameters for wheeled robots.

After SLAM methods started to become more prevalent, it became possible to use probability distributions reported from the SLAM algorithm to estimate a robot's ground-truth pose. This estimate could then be used in conjunction with calibration methods to automatically learn the parameters of a motion model by simply moving the robot.

Roy and Thrun [4], for example, use a maximum likelihood method for estimating model parameters. The parameters $\vec{\theta}$ are updated with an exponential estimator that integrates the parameter values $\vec{\theta}^*$ that maximize the likelihood function $p(\vec{s}_{t+1} \mid \vec{s}_t, \vec{o}, \vec{\theta}^*)$, where $\vec{s}_t$ are laser-scan measurements and $\vec{o}$ are odometry measurements. By using an exponential estimator, incremental updates to the parameters are possible which do not depend on keeping a history of sensor-measurement data.

In work done by Eliazar and Parr [5], the model parameters are learned using an Expectation-Maximization (EM) method, in which the expectation step involves using the SLAM algorithm (in this case, DP-SLAM 2.0 [6]) to generate a set of possible trajectories for a given set of motion model parameters and associated likelihoods, and the maximization step uses a least-squares approach for determining the set of model parameters that maximizes the likelihood values.

While the least-squares method presented in [5] assumes that a history of the training data is available, Visatemongkolchai and Zhang [7] apply two incremental least-squares methods in order to learn model parameters. Because of their incremental nature, these methods can be used in an on-line fashion, updating the model parameters after each new set of measurements.

Stronger and Stone [8] introduce a technique for calibrating sensor and motion models simultaneously. These models are represented in a deterministic manner using polynomials. The polynomial coefficients are learned with a two-step cyclic algorithm in which the first step estimates the sensor-model parameters given the current motion-model, and the second step estimates the motion-model parameters given the current sensor-model.

The method presented by Kaboli et al. [9] also calibrates sensor and motion models simultaneously, but in contrast to [8], uses a Markov chain Monte Carlo (MCMC) technique in which samples are drawn from a posterior distribution over model parameters. To estimate the true model parameters, these samples are either averaged, or the maximum a posteriori sample is selected. Additionally, the samples can be used to approximate model posteriors. Each of these methods were tested in a Monte Carlo Localization scenario on a simulated wheeled robot and on a Sony AIBO robot.

Martinelli et al. [10] present a method of estimating the parameters of an odometry error model by using a modified Extended Kalman Filter (EKF) to simultaneously estimate the robot pose and error parameters. The odometry model relates odometry measurements (i.e. wheel encoder measurements) of a wheeled robot to the robot's estimated true odometry. The error model is split into systematic and non-systematic components, and each component is separately estimated based on the other component's estimate.

Sjoberg et al. [11] present a method that uses a slightly simplified deterministic model from that presented in [5], but in which the model's random variables are represented using a bimodal Gaussian Mixture Model. The method depends on the use of a SLAM method (DP-SLAM 2.0 is used), and the model parameters are estimated by observing the motion of the particles in a particle-filter. As in [4], an exponential estimator was used to make the method work in an online fashion, in which the decay factor is adjusted according to the average quality of the particles in a particle-set.

The work done by Hoffmann [12] does not deal with the problem of parameter estimation, but instead presents a way of incorporating proprioceptive information into the motion model. To do this, the motion model's error is decomposed into two components: the error $\epsilon_{coll}$ due to collisions and slippage, and the error $\epsilon_{odo}$ intrinsic to the robot morphology and odometry sensors. Collisions and slippage conditions are modeled as the states of a state-machine, and the value of $\epsilon_{coll}$ depends on which state the robot finds itself in. Transitions between states occur due to the observation of specific proprioceptive data patterns.

*Suitability for Legged Robots*

Nearly all of the methods in the previously mentioned works define the motion model as a set of fixed form "ideal model"[1] equations to which uncertainty is added by treating some of the variables as random. Furthermore, every one of these methods requires that the motion model be parameterized with a fixed number of parameters.

For kinematically complex legged robots, developing the equations of an ideal model can be very difficult and time consuming.[2] Because of this, *these methods generally do not lend themselves well to modelling the motion of legged robots*.

To properly capture the motion model of such a legged robot, as few as possible assumptions should be made about the form of the motion model. For this reason, we have adopted the representation presented in the next section, which provides a flexible form that can be dynamically changed over time.

III. MOTION MODEL REPRESENTATION

*A. Introduction*

Motion models are frequently defined in the form of a conditional state-transition probability density, where the state is defined as the robot's pose, and the probability is conditioned on the commands issued to the robot. In unobstructed areas, the change in a robot's pose from the current state to the next state depends much more on the command issued to the robot than it does on the robot's current pose. Because of this, the motion model can be simplified by representing it as a probability of the *change in pose*[3] conditioned upon the command issued.

---

[1] by "ideal model", we mean a model which assumes an ideal, non-stochastic world.

[2] There are some legged robots (e.g. the Sony AIBO) whose kinematics are simple enough to develop equations approximating an ideal model.

[3] This change in pose is measured relative to the robot's pose prior to the execution of a command.

We have chosen to represent the motion model using a *dynamic* mixture-of-Gaussians model, in which the number of Gaussian components can vary in order to best fit the characteristics of the underlying system. This weighted set of Gaussian components represents, for each unique command, a probability distribution over the change in pose of the robot.

*B. Formal Description*

Formally, we can represent the motion model as $\mathcal{M} : C \times E \to P$ where $C$ is the set of all commands, $E$ is the set of environmental states[4], and $P$ is the set of probability density functions $p(\vec{x})$, where $\vec{x}$ is a change in pose measurement. In simple cases where the robot's environment is not expected to change much, $E$ may be treated as containing only a single static state. In this case, the model can be written as $\mathcal{M}_S : C \to P$. In practice, a robot's command space (i. e. the range of possible commands that can be issued to the robot) can be discretized into $n$ commands $\{c_1, c_2, ..., c_n\}$, where each command is a vector representing various dimensions that can be controlled for a robot (e. g. for a wheeled robot, these dimensions might be the angular velocity of each wheel)[5]. If the commands are discretized in this way, then according to the mapping defined by $\mathcal{M}_S$, a corresponding probability density in $P$ (that is a likelihood function for $\vec{x}$) exists for each of these commands. We represent each command $c_j$'s corresponding density function $p(\vec{x} \mid c_j) \in P$, as a variable-sized set of "weighted Gaussian" pairs,

$$G_j \equiv \{(g_{j1}(\vec{x}), w_{j1}), (g_{j2}(\vec{x}), w_{j2}), \ldots, (g_{jm}(\vec{x}), w_{jm})\},$$

such that

$$p(\vec{x} \mid c_j) = \sum_{i=1}^{m} \hat{w}_{ji} g_{ji}(\vec{x}), \tag{1}$$

where $g_{ji}(\vec{x})$ is a conditional multivariate Gaussian distribution:

$$g_{ji}(\vec{x}) = p_{ji}(\vec{x} \mid c_j) \sim \mathcal{N}(\mu_{ji}, \Sigma_{ji}), \tag{2}$$

and

$$\hat{w}_{ji} = w_{ji} \,/ \sum_{k=1}^{m} w_{jk}. \tag{3}$$

A schematic depiction of this representation for the case of discrete command spaces can be seen in Figure 1. The $w_{ji}$ values are *unnormalized* weights, and are equivalent to the number of $\vec{x}$ samples that have contributed to the corresponding Gaussian (more on this in the next section).

It is of course possible to fix the quantity $m$ of the components representing $p(\vec{x} \mid c_j)$ in (1), which would result in a standard Gaussian Mixture Model (GMM) representation. By allowing $m$ to vary, it is possible for the motion model to capture more precisely the nature of $p(\vec{x} \mid c_j)$ (similar to the behavior of a particle filter whose number of particles is allowed to vary), but then we cannot depend on the same methods traditionally used with GMMs. We call this flexible representation a *Dynamic Gaussian Mixture Model* (DGMM), and present a method in the following section that can handle this relaxed constraint.

---

[4]An environmental state simply represents some information about the environment immediately surrounding the robot. For example, it could include information about the terrain's roughness, slope, or moisture content.

[5]Although the remainder of this paper assumes that the command space is discrete, the representation presented here can also be used with continuous command spaces.

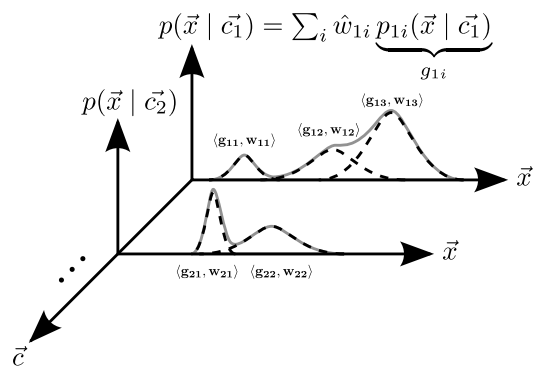

Fig. 1. Motion model representation: for each command in the discretized command space, a conditional probability distribution is maintained as a weighted sum of Gaussians.

## IV. UPDATE METHOD

In this section we describe the method we use for updating the parameters and structure of a DGMM based motion-model. The method can easily incorporate multi-modal sensory data into the motion-model, but in this discussion, we restrict the measurements to include only the robot's pose for the sake of simplicity. The primary algorithm of this method is depicted in Algorithm 1.

---

**Algorithm 1**: Incremental Motion Model Update

**Data**: Sequence of $\langle command, pose\ measurement \rangle$ pairs.
**Result**: Dynamically updated motion-model for each *command*.
Select and perform command $c_j \in C$ on robot;
$\zeta \leftarrow$ new pose measurement;
**repeat**
  $\zeta_{previous} \leftarrow \zeta$;
  Select and perform command $c_j \in C$ on robot;
  $\zeta \leftarrow$ new pose measurement;
  $\vec{x} \leftarrow \zeta - \zeta_{previous}$; // *change in pose*
  ADDSAMPLE$(c_j, \vec{x})$ ; // *update* $p(\vec{x} \mid c_j)$
**end**

---

Initially, the robot's pose $\zeta_0$ is recorded, and a command $c_j$ is chosen from the set of possible commands $C$. This command is issued to the robot, and the final pose $\zeta_1$ is recorded after the command has completed its execution. The measurement vector $\vec{x}$ is calculated as $\zeta_1 - \zeta_0$ (i. e. the robot's change in pose), and incorporated into the motion model via the ADDSAMPLE algorithm (described below). Again, a command is selected, and this procedure is repeated ad infinitum, calculating the measurement vector in the same manner (subtracting the previous pose measurement from the current).

---

**Algorithm 2**: ADDSAMPLE: Incorporate a differential sample measurement into the model.

**Data**: One $\langle command, \vec{x} \rangle$ pair.
**Result**: Updated density function for $command$.
$r \leftarrow U(0, 1)$; *// sample from uniform pdf*
*// normalized N-dimensional density from current*
*// model with normalization factor $W$*
$d \leftarrow W\, p(\vec{x} \mid command)$;
$n \leftarrow$ size of sample-set used to form $p(\vec{x} \mid command)$;
$k \leftarrow$ merge likelihood constant;
*// calculate merge threshold value*
$t \leftarrow 1 - (1 - d) \exp\{-kn\}$;
**if** $r < t$ **then**
| Merge $\vec{x}$ into an existing Gaussian;
**else**
| Add $\vec{x}$ as a new Gaussian;
**end**

---

The ADDSAMPLE algorithm starts out with a command $c_j$ in $C$, and an associated pose difference vector $\vec{x}$. A decision is probabilistically made to either merge $\vec{x}$ into an existing Gaussian in the model, or to add a new Gaussian to the model whose mean is $\vec{x}$. Which option is chosen depends on two main factors:

1) $d = W\, p(\vec{x} \mid c_j)$, a normalized[6] likelihood of the N-dimensional vector $\vec{x}$ given $c_j$, and
2) $n$, the number of samples already incorporated into $p(\vec{x} \mid c_j)$.

These values are used to calculate a *merge threshold value*

$$t = 1 - (1 - d)e^{-kn} \qquad (4)$$

which takes on a value between 0 and 1. As $d \to 1$ and $n \to \infty$, $t$ approaches 1, which means that it is more likely that $\vec{x}$ will be used to update the parameters of an existing Gaussian. As $t$ approaches 0 (i.e. $d \to 0$ and $n \to 0$), it becomes more likely that a new weighted Gaussian will be added to $G_j$. The $k$ factor in (4) is used to adjust the rate at which the merge threshold transitions towards 1 (i.e. towards the state in which samples will always be merged). This factor can be automatically chosen by using, for example, techniques based on cross-validation.

If a new weighted Gaussian based on $\vec{x}$ is added to $G_j$, it is assigned a mean of $\vec{x}$ and an identity covariance matrix, and its mixture parameter (i.e. weight) is initialized to 1. On the other hand, if $\vec{x}$ is to be incorporated into an existing Gaussian, the Gaussian is chosen probabilistically such that the probability of choosing $g_{ji}$ from $G_j$ is proportional to $g_{ji}$'s normalized density multiplied by $w_{ji}$, and evaluated at $\vec{x}$. After $g_{ji}$ has been selected, we can determine a set of update equations to incorporate $\vec{x}$ as follows. Given the mean vector and covariance matrix of an unbiased estimator of $n$ samples,

$$\bar{y}_n = \frac{1}{n} \sum_{i=1}^{n} y_i, \qquad (5)$$

$$\Sigma_n = \frac{1}{n-1}(\sum_{i=1}^{n} y_i y_i^T) - \frac{n}{n-1} \bar{y}_n \bar{y}_n^T, \qquad (6)$$

where $y_1, y_2, \cdots, y_n$ are the samples that have contributed to $g_{ji}$ prior to incorporating the new sample $y_{n+1}$, our goal is to determine the updated mean and covariance,

$$\bar{y}_{n+1} = \frac{1}{n+1} \sum_{i=1}^{n+1} y_i, \qquad (7)$$

$$\Sigma_{n+1} = \frac{1}{n} \left\{ (\sum_{i=1}^{n+1} y_i y_i^T) - \bar{y}_{n+1} \bar{y}_{n+1}^T \right\}. \qquad (8)$$

---

[6]The density function is normalized such that its maximum value is 1.

After some algebraic manipulation of the above equations, the new mean and covariance matrix can be written as

$$\bar{y}_{n+1} = \frac{n}{n+1} \bar{y}_n + \frac{1}{n+1} y_{n+1}, \qquad (9)$$

$$\Sigma_{n+1} = \frac{n-1}{n} \Sigma_n + \bar{y}_n \bar{y}_n^T + \frac{1}{n} y_{n+1} y_{n+1}^T \\ - \frac{n+1}{n} \bar{y}_{n+1} \bar{y}_{n+1}^T. \qquad (10)$$

Finally, because the unnormalized weight $w$ of a Gaussian equals the number of samples which have contributed to it, these update equations can be written in terms of the previously used notation as:

$$w_{new} = w_{old} + 1 \qquad (11)$$

$$\vec{\mu}_{new} = \frac{w_{old}}{w_{new}} \vec{\mu}_{old} + \frac{1}{w_{new}} \vec{x} \qquad (12)$$

$$\Sigma_{new} = \frac{w_{old} - 1}{w_{old}} \Sigma_{old} + \vec{\mu}_{old} \vec{\mu}_{old}^T \\ + \frac{1}{w_{old}} \vec{x} \vec{x}^T - \frac{w_{new}}{w_{old}} \vec{\mu}_{new} \vec{\mu}_{new}^T \qquad (13)$$

Note that these equations only apply when adding a measurement to an already-existing Gaussian (i.e. when $w_{old} > 0$). Because these update equations are incremental in nature, the model can be updated without the need to store a history of samples. This incremental nature reduces both the memory and computational requirements of this algorithm considerably, in contrast to methods in which some or all samples are required to re-estimate the density.

## V. INCORPORATING TERRAIN DATA

The discussion thus far has assumed that the model represents only a simple mapping between commands and pose differences. The DGMM based model can, however, be extended to include exteroceptive and proprioceptive data[7].

---

[7]Data is most effectively incorporated when it can in some way be represented as a vector in a Euclidean space. Fortunately there is often a means of representing data in this way.

## A. Method of Incorporating Additional Data

Specifically, if we have some data (e.g. information about the terrain) we would like to incorporate into the model, represented as $\vec{z}$, the change in pose vector $\vec{x}$ can be augmented by $\vec{z}$ to form a new data vector

$$\vec{d} = \vec{x} \| \vec{z} \\ = \langle x_1, \cdots, x_n, z_1, \cdots, z_m \rangle. \tag{14}$$

It is then possible to use $\vec{d}$ in the same way that $\vec{x}$ was previously used in Section IV. The only necessary change is to modify Algorithm 1 such that the measurement $\vec{z}$ is taken before the command $c_j$ is issued to the robot. The vector $\vec{d}$ is then formed and passed to ADDSAMPLE in place of $\vec{x}$.

The resulting motion model will now represent $p(\vec{x} \| \vec{z} \mid c_j)$, where each Gaussian component $g_{ji}$ of this density function has a mean vector $\mu_{ji}^d = \mu_{ji}^x \| \mu_{ji}^z$ and covariance matrix $\Sigma_{ji}^d = \begin{bmatrix} \Sigma_{ji}^{xx} & \Sigma_{ji}^{xz} \\ \Sigma_{ji}^{zx} & \Sigma_{ji}^{zz} \end{bmatrix}$.

## B. Using the Enhanced Motion Model

After building a motion model with these augmented sample vectors, one typically wants to know just how the model can be used. For the purposes of modelling motion, the density $p(\vec{x} \| \vec{z} \mid c_j)$ is not directly useful. Rather, we would like to have $p(\vec{x} \mid c_j, \vec{z})$. To calculate the latter density from the former, we begin by writing the conditional probability relationship for the latter density:

$$p(\vec{x} \mid c_j, \vec{z}) = \frac{p(\vec{x} \| \vec{z} \mid c_j)}{p(\vec{z} \mid c_j)}. \tag{15}$$

Because $p(\vec{x} \| \vec{z} \mid c_j)$ is represented as a mixture of Gaussians, the numerator in (15) can be written as

$$p(\vec{x} \| \vec{z} \mid c_j) = \sum_{i=1}^{m} \hat{w}_{ji}^d g_{ji}^d(\vec{x} \| \vec{z}), \tag{16}$$

where

$$g_{ji}^d(\vec{x} \| \vec{z}) = p(\vec{x} \| \vec{z} \mid c_j) \sim \mathcal{N}(\mu_{ji}^d, \Sigma_{ji}^d). \tag{17}$$

Using the properties of Gaussian distributions, the denominator in (15) can be written as

$$p(\vec{z} \mid c_j) = \sum_{i=1}^{m} \hat{w}_{ji}^z g_{ji}^z(\vec{z}), \tag{18}$$

where $\hat{w}_{ji}^z = \hat{w}_{ji}^d$ and

$$g_{ji}^z(\vec{z}) = p(\vec{z} \mid c_j) \sim \mathcal{N}(\mu_{ji}^z, \Sigma_{ji}^{zz}). \tag{19}$$

Thus, by using (16) and (18), both which can easily be calculated from our motion model, we can determine $p(\vec{x} \mid c_j, \vec{z})$, which represents *an improved motion model that takes advantage of additional sensory data.*

## VI. MODEL EVALUATION

The performance of our DGMM based motion modelling technique was evaluated in terms of (a) its ability to represent unknown distributions, and (b) the extent to which incorporating terrain data as described in the previous section improved the model.

## A. Representational Capability

*a) Effect of merge likelihood constant:* The representational capability of the DGMM, when used with the algorithms described in Section IV, depends on the merge likelihood constant $k$ in (4). This constant controls the complexity of the resulting model, influencing the number of Gaussians in a trained model[8]. It is therefore important to choose a good value for $k$ when updating the motion model. To see the effect that $k$ has on the number of Gaussians in a model, a dataset was generated from a known distribution, and for several values of $k$, this dataset's density was estimated. The relationship between $k$ and the final number of Gaussians in the estimate is shown in Figure 2.

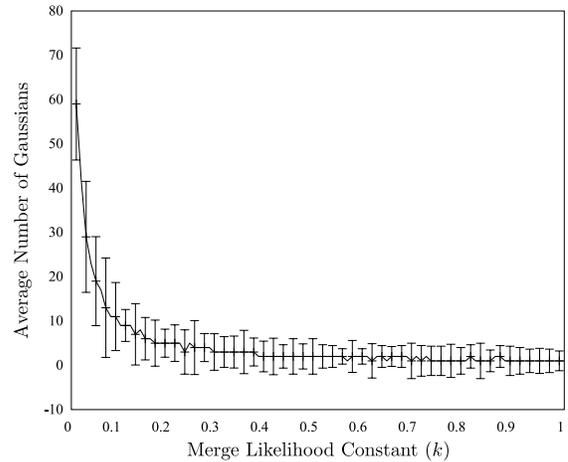

Fig. 2. Effect of varying the merge likelihood constant: as $k$ increases, the number of Gaussians in the learned model decreases (for a given sample distribution).

One method of choosing $k$ is to collect a large set of samples, and perform a cross-validation on this dataset, examining the effect that $k$ has on the likelihood of a validation set given a model trained with a training set. The $k$ value that maximizes this likelihood can be selected, and used for further updates to the model. The initial corpus of samples can be collected from either a real or a simulated robot. Provided that the simulation is accurate enough, using a simulation has the advantage of allowing a large number of samples to be collected, so that a good $k$ value can be chosen prior to using the motion model with the real robot.

---

[8]The number of Gaussians in a model depends not only on $k$, but also on the dataset the algorithm is applied to. Nonetheless, an appropriate $k$ generally results in a good density estimate despite variations in a dataset's optimal number of Gaussians.

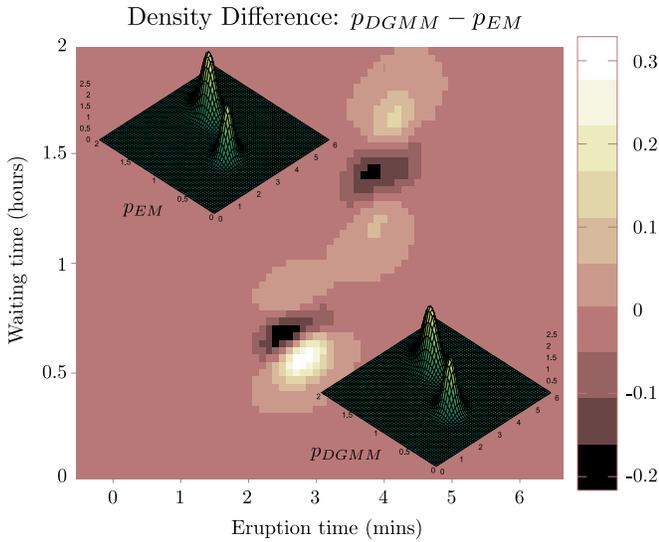

Fig. 3. Comparison of density estimate $p_{EM}$ derived using the EM algorithm (upper left), and the density estimate $p_{DGMM}$ derived using the online DGMM method (lower right), on the Old Faithful dataset. The heat map in the background shows the difference between $p_{DGMM}$ and $p_{EM}$.

*b) Comparison with Expectation Maximization:* In order to test the quality of our method, it was compared with an Expectation-Maximization (EM) based approach in its ability to model the "Old Faithful" benchmark dataset. The EM algorithm was used to estimate the parameters of a 2-Gaussian GMM in an offline manner (i. e. using all of the sample-data in each iteration). Just as it was necessary to select in advance the number of Gaussians for the EM approach, $k = 0.7$ was chosen for the online update algorithm, resulting in density estimates of this dataset that most often contain two Gaussians. The resulting density estimates are shown in Figure 3. The upper-left and lower-right plots show the resulting density estimate from the EM and online DGMM methods, respectively. Though barely perceptible from these plots, the difference between the plots is shown in the heat map in the background. These results demonstrate that the DGMM online method can produce density estimates comparable to the EM method. It should be noted that because the DGMM online method is stochastic, the estimates it produces can vary depending on the random seeds used. To get a better idea of the method's general performance, it was allowed to run until 100 distributions containing 2 Gaussians had been produced. The Mean Integrated Square Error (MISE) between the each of these 100 distributions and the EM generated distribution were calculated, resulting in an average MISE of 0.0358 with a standard deviation of 0.0317. It is clear from these results that the density estimates generated by these two methods are very similar, though the computational and storage requirements are quite different, since the online DGMM method requires only a single-pass and does not need to store the samples.

## B. Effect of Incorporating Terrain Data

The cross-validation method previously mentioned in the context of selecting $k$ can also be used to evaluate the accuracy of a motion model: a set of samples (either via simulation or from a real robot) is taken, and one portion of these samples is used for building the model, while the remaining samples are used for validating the model's ability to predict motion.

We chose to use this technique to assess the extent to which terrain information contributes to model accuracy. In particular, a total of 390 data samples from the SCORPION robot [13] were collected, in which every possible command of the form $\langle long, lat, turn \rangle$ was issued 5 times with three different initial robot orientations (i.,e. each command is issued a total of 15 times), where $long, lat, turn \in [-0.5, 0, +0.5]$[9]. Not including the *no-op* command $\langle 0, 0, 0 \rangle$, this results in a set of 26 distinct commands. With the experimental setup pictured in Figure 4, the following procedure was used for collecting the data:

1) A command $c_j$ is chosen randomly (without replacement) from the above defined command set.
2) The robot is placed on an 18° inclined slope with a starting orientation $\theta \in \{0°, 90°, -90°\}$[10]. The orientation angle is cycled each time this step is performed.
3) The current robot pitch and roll angles are recorded in the vector $\vec{z}$. This serves (for the case where the terrain is an inclined plane, as in this experiment) as an indirect measurement of the shape of the terrain the robot is on.
4) The robot is issued the command $c_j$. Each command's execution requires a fixed time duration to carry out.
5) Before and after executing a command, the robot's pose is recorded using a marker based visual pose tracking system[11].
6) The sample $(c_j, \vec{z} \| \vec{x})$ is added to the sample set $S_1$, where $\vec{z} = \langle robot\ pitch, robot\ roll \rangle$ and $\vec{x}$ is the change in pose $\langle \Delta x, \Delta y, \Delta z, \Delta roll, \Delta pitch, \Delta yaw \rangle$.
7) Steps 3-6 are repeated 5 times.
8) Steps 2-7 are repeated until the robot has been placed with each starting orientation.
9) Steps 1-8 are repeated until all commands have been chosen.

The samples from this experiment were used for cross-validation in which the model was learned using randomly selected training sets, and was validated by calculating the log-likelihood of the training data given the learned model, $L = \sum_{n=1}^{N} \ln \left\{ \sum_{i=1}^{M} \hat{w}_{ji} g_{ji}(\vec{x}_n) \right\}$, where $\hat{w}_{ji}$ is the associated normalized mixture-parameter of the $i^{\text{th}}$ Gaussian in the model for a command $c_j$.

[9]The $\langle long, lat, turn \rangle$ values influence the robot's motion in the *forward-backward*, *lateral-left-right*, and *turning-left-right* directions, respectively.
[10]When $\theta = 90°$, the robot is turned so that it looks down the incline.
[11]Note that although we use an external motion tracking system, samples can also be acquired from other sources such as SLAM or GPS.

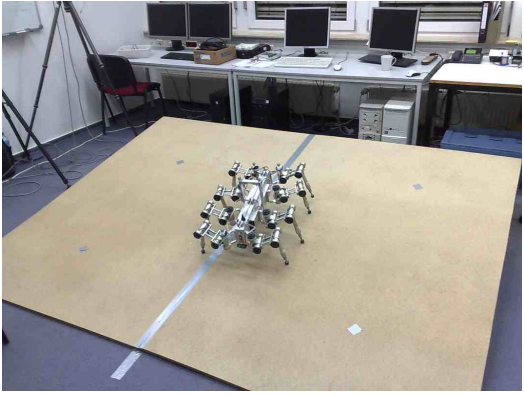

Fig. 4. Experimental Setup: the Scorpion robot executes several movement commands in different orientations on a sloped plane.

TABLE I
MODEL QUALITY COMPARISON: A 10-FOLD CROSS-VALIDATION IS PERFORMED FOR DATA SETS WITH AND WITHOUT TERRAIN INFORMATION. SHOWN ARE THE MEAN AND STANDARD-DEVIATION OF THE LOG LIKELIHOOD VALUES, AVERAGED OVER 10 RUNS.

| Data Set | # Samples | Log Likelihood | |
|---|---|---|---|
| | | Without Terrain | With Terrain |
| Real | 390 | $-371.9 \pm 3.9$ | $-162.2 \pm 4.0$ |
| Simulated | 830 | $308.5 \pm 7.5$ | $402.5 \pm 7.9$ |

The cross validation was performed using the collected data for two different cases:

*c) Case 1:* The first case in which we performed cross validation uses the "full perception" sample set $S_1$ that contains all the measured information about the terrain. In this case, the conditional density in (15) is calculated from the model.

*d) Case 2:* In the second case, the model is trained and validated using a "limited perception" sample set $S_2$, in which the $\vec{z}\|\vec{x}$ component of each sample $S_1$ is replaced with $\vec{x}$. In other words, for each sample $(c_j, \vec{z}\|\vec{x})$ in $S_1$, there is a corresponding sample $(c_j, \vec{x})$ in $S_2$. In this case, the model estimates the conditional densities in (1).

For both cases, a stratified 10-fold cross validation was performed 10 times, and the results averaged. The results for both cases are shown in Table I. To test the validity of this comparison when there are a larger number of samples, we generated another data set (containing a total of 830 samples) using a simulation. This data was analyzed in exactly the same manner as the data collected from the real robot, and the results are also shown in Table I. With both the real and simulated data, the likelihood was much higher in the case where terrain data was included in the model. This shows that *incorporating and making use of terrain data in the model substantially improves the extent to which the model describes the data.*

## VII. CONCLUSIONS AND FUTURE WORK

In this paper we have introduced a method for representing and capturing motion models which (a) alleviates the need to hand-design the form of a deterministic motion model, and (b) provides a straightforward means of incorporating exteroceptive and proprioceptive data into the motion model. Furthermore, it has been demonstrated experimentally that incorporating terrain information into the model improves the model's accuracy.

In the future we plan to investigate the scalability of DGMM based models with respect to incorporating large amounts of environmental data. Additionally, researching an effective means of adjusting the DGMM algorithm's $k$ parameter in an online manner would improve the adaptability of the algorithm to nonstationary processes. Furthermore, the basic online algorithm as presented here assumes that the samples it receives are not temporally correlated. We are presently investigating ways of making the algorithm perform well even when supplied with a temporally correlated sequence of samples. Finally, we plan to extend preliminary work done on smoothing algorithms that operate on learned DGMM models.

ACKNOWLEDGMENT

This work was supported by the German Science Foundation (DFG) under contract number SFB/TR-8 (A3). Special thanks to Thijs Jeffry de Haas, Tchando Kongue, and Daniel Beßler for their assistance with the experiments.